%% file: main.tex
\documentclass[acmtog]{acmart}
\acmSubmissionID{papers\_1234}

\input{sections/_macros}

\usepackage[ruled]{algorithm2e} 

\newcommand{\methodname}{\textsc{SAM3D-Part}}
\AtBeginDocument{%
  }

\copyrightyear{2026}
\acmYear{2026}
\setcopyright{cc}
\setcctype{by}
\acmConference[SA Conference Papers '26]{SIGGRAPH Asia 2026 Conference Papers}{December 01--04, 2026}{Kuala Lumpur, Malaysia}
\acmBooktitle{SIGGRAPH Asia 2026 Conference Papers (SA Conference Papers '26), December 01--04, 2026, Kuala Lumpur, Malaysia}
\acmDOI{10.1145/3829340.3842187}
\acmISBN{979-8-4007-2842-6/2026/12}

\begin{document}

\title{\methodname{}: Interactive Part Selection and Generation from 3D Objects}

\author{Jiahao Chang}
\affiliation{%
  \institution{SSE, CUHKSZ and FNii-Shenzhen and Meshy AI}
  \country{China}
}
\email{224010128@link.cuhk.edu.cn}
\authornote{Work done during internship at Meshy AI.}
\authornote{These authors contributed equally to this work.}

\author{Dong Du}
\affiliation{%
  \institution{Nanjing University of Science and Technology}
  \country{China}
}
\email{dongdu@njust.edu.cn}
\authornotemark[2]

\author{Wanhu Sun}
\affiliation{%
  \institution{SSE, CUHKSZ}
  \country{China}
}
\email{wanhusun@gmail.com}

\author{Yujian Zheng}
\affiliation{%
  \institution{MBZUAI}
  \country{United Arab Emirates}
}
\email{yujian.zheng@mbzuai.ac.ae}

\author{Chuanyu Pan}
\affiliation{%
  \institution{Meshy AI}
  \country{USA}
}
\email{chuanyupan@meshy.ai}
\authornote{Project leader.}

\author{Bowen Zhao}
\affiliation{%
  \institution{Meshy AI}
  \country{USA}
}
\email{bowen@meshy.ai}

\author{Chongjie Ye}
\affiliation{%
  \institution{SSE, CUHKSZ and FNii-Shenzhen}
  \country{China}
}
\email{chongjieye@link.cuhk.edu.cn}

\author{Yuanming Hu}
\affiliation{%
  \institution{Meshy AI}
  \country{USA}
}
\email{yuanmhu@gmail.com}

\author{Xiaoguang Han}
\affiliation{%
  \institution{SSE, CUHKSZ and FNii-Shenzhen and GenuX}
  \country{China}
}
\email{hanxiaoguang@cuhk.edu.cn}
\authornote{Corresponding authors.}

\renewcommand{\shortauthors}{Chang et al.}

\input{sections/0_abstract}


\begin{teaserfigure}
  \centering
  \includegraphics[width=1.\linewidth]{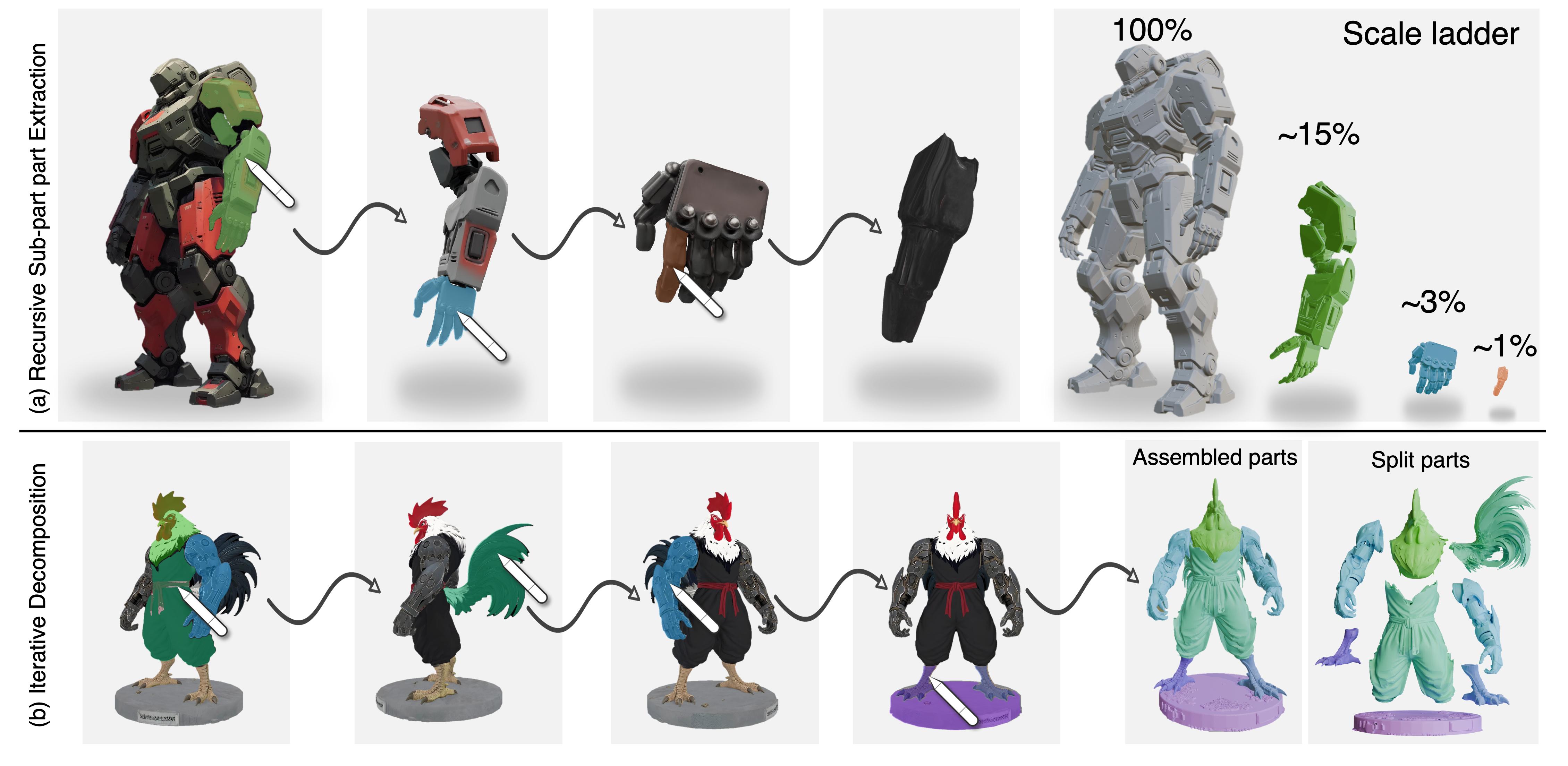}
  \caption{Given an object mesh, SAM3D-Part enables interactive part selection on a rendered view and generates complete 3D part meshes. It supports fine-grained, controllable selection (a) and full object decomposition (b) from arbitrary viewpoints.}
  \label{fig:teaser}
\end{teaserfigure}


\maketitle

\input{sections/1_introduction}
\input{sections/2_relatedwork}
\input{sections/3_method}
\input{sections/4_experiment}
\input{sections/5_conclusion}

\begin{acks}
This work was supported in part by the Guangdong S\&T Programme under Grant No.~2024B0101030002; the Basic Research Project of the Hetao Shenzhen--Hong Kong S\&T Cooperation Zone under Grant No.~HZQB-KCZYZ-2021067; the National Natural Science Foundation of China (NSFC) under Grant No.~62293482; the Guangdong Provincial Fund for Distinguished Young Scholars under Grant No.~2023B1515020055; the Shenzhen Outstanding Talents Training Fund under Grant No.~202002; the Guangdong Provincial Key Laboratory of Future Networks of Intelligence under Grant No.~2022B1212010001; the Shenzhen Key Laboratory of Big Data and Artificial Intelligence under Grant No.~SYSPG20241211173853027; the Guangdong Province Radio Science Data Center under Grant No.~2025B1212070001; the National Key R\&D Program of China under Grant No.~2018YFB1800800; the National Natural Science Foundation of China (No. 62502209); and the Fundamental Research Funds for the Central Universities (No. 30925010538).
\end{acks}

\clearpage
\bibliographystyle{ACM-Reference-Format}
\bibliography{reference}

\clearpage
\input{sections/6_figs_only}


\end{document}

%% file: sections/_macros.tex
\usepackage{graphicx}
\usepackage{amsmath}

\usepackage{amssymb}

\usepackage{xcolor}
\usepackage[normalem]{ulem}
\usepackage{enumitem}
\usepackage{xspace}
\usepackage{booktabs}
\usepackage{colortbl}
\usepackage{multirow}
\usepackage{makecell}
\usepackage{gensymb} 
\usepackage{mathtools}

\usepackage[labelsep=period]{caption}
\definecolor{MyDarkRed}{rgb}{0.46, 0.16, 0.16}
\definecolor{MyDarkBlue}{rgb}{0.16, 0.16, 0.66}



%% file: sections/0_abstract.tex
\begin{abstract}

Part-level control is essential for modern 3D asset creation, where objects are frequently edited, reused, animated, or fabricated through their individual components. In many such workflows, users need only several specific components rather than a complete object decomposition. However, existing 3D generation methods produce all parts regardless of user intent, while promptable 3D segmentation methods typically output partial surfaces instead of reusable complete meshes. In addition, image-conditioned part generators further struggle to preserve hidden geometry and accurate placement without directly conditioning on the source mesh.
To address these problems, we present SAM3D-Part, a prompt-driven framework for selective part generation from input 3D object meshes. Given a source mesh and a part prompt, SAM3D-Part first encodes the source geometry into compact mesh features and aligns them with the rendered image, selective mask, and point-map observations via pixel-wise channel fusion. The fused representation conditions a feed-forward generative model to produce only the queried component as a completed mesh. To place the generated part back into the source coordinate frame, SAM3D-Part predicts dense per-voxel correspondences and estimates the part transformation from distributed spatial evidence rather than a single global pose code. For sequential multi-part queries, previously generated parts are stored in a part cache and reused as contextual constraints, reducing conflicts among independently requested components.
Extensive experiments and ablations demonstrate that SAM3D-Part can significantly improve source alignment, reduce conditioning cost, and enable consistent selective part generation, achieving state-of-the-art.
Code and weights will be available at \url{https://github.com/Jiahao620/sam3d-part}.
\end{abstract}




%% file: sections/1_introduction.tex
\section{Introduction}
\label{sec:intro}

With the rapid development of 3D content creation, part-aware control is becoming a central requirement in various applications. Downstream workflows such as rigging, animation, part replacement, kitbashing, and 3D printing often require individual components to be isolated, edited, or reused independently from the full object. In these workflows, users rarely need an exhaustive decomposition of the entire asset. For instance, a modeler may want to extract only a dragon’s wing, replace only a chair leg, or print only a mechanical connector. This selective interaction mode is already familiar in 2D image editing, where promptable segmentation enables users to specify exactly the region they care about. For 3D asset creation, we seek an analogous design that takes an existing object mesh and a user-specified part prompt as input, then returns a reusable part mesh geometrically aligned with the input object.

However, existing part-level 3D generation methods~\cite{liu2024part123,chen2025partgen,yang2025omnipart,zhang2025bang,he2025unipart} mainly focus on fully automatic object decomposition, aiming to recover all constituent parts of an object. When users need only one or two components, generating all parts is computationally inefficient and still requires additional selection from the resulting decomposition. 
More importantly, these methods cannot ensure precise control over partition granularity.
Although promptable 3D segmentation methods~\cite{yang2024sampart3d,ma2025p3,li2026segvigen} provide a more selective interface, they typically output incomplete meshes using surface masks rather than complete part meshes suitable for downstream reuse. 
Recent image-to-part generation methods, especially SAM3D~\cite{chen2025sam}, are closer to selective part generation, but they condition on rendered views instead of the input mesh. This limits their ability to produce source-aligned part meshes, especially for parts with occluded structures, ambiguous boundaries, or strong dependence on the surrounding geometry.

In this paper, we address prompt-driven selective part extraction from existing object meshes and introduce an effective framework, named SAM3D-Part. Given an input mesh, we first select a suitable viewpoint, such as the front view, and render the mesh into a 2D image. The user then specifies the desired component through simple point interactions, using one or a few clicks on the rendered view. Based on these point prompts, we apply a SAM-based segmentor~\cite{ravi2025sam} to obtain the corresponding 2D part mask, which serves as an intuitive prompt for subsequent 3D part generation. Unlike surface segmentation methods that output only labels or masks, SAM3D-Part formulates the task as conditional 3D generation and directly produces a complete part mesh.

To ensure that the generated part remains consistent with the source object, SAM3D-Part conditions the generation process on both image-space prompts and mesh-space geometry. Specifically, the rendered RGB image, the selected part mask, and the point-map observation are encoded as 2D prompt tokens, where appearance, user attention, and spatial information are fused at the same image-patch locations. In parallel, the input mesh is encoded into 3D geometry tokens using a pretrained shape encoder. 
These 2D and 3D condition tokens are then combined along the token dimension and injected into the generative model through cross-attention. This design provides the generator with both the user's part-level intent and the complete geometric context of the original mesh, including regions that may be invisible from the selected view. 

Built upon a two-stage sparse 3D generation framework, SAM3D-Part first predicts the coarse occupancy of the queried part and a dense XYZ correspondence map from the generated part to the source mesh frame. Instead of regressing a single global pose, the dense correspondence provides voxel-level spatial evidence, from which the scale and translation of the generated part can be estimated in closed form. 
The second stage then refines the predicted structure into high-resolution geometry and decodes it into a standalone mesh. For sequential multi-part queries, we further maintain a part cache that records previously generated components and attaches this history to the mesh tokens, helping later queries avoid conflicts and maintain consistency across extracted parts.

We evaluate SAM3D-Part on multiple object categories under both single-part and multi-part settings. Extensive experiments and ablation studies are conducted to assess part quality, source alignment, and consistency across sequential queries. 
Experimental results show that SAM3D-Part consistently outperforms existing baselines in selective part generation, achieves better alignment with the input object, and maintains stable performance when generating multiple parts sequentially.

Our contributions are summarized as follows:
\begin{itemize}[left=0pt]
    \item We formulate prompt-driven selective part extraction from existing 3D meshes and propose a novel framework, i.e., SAM3D-Part, aiming to generate reusable and source-aligned meshes for only the user-specified components.

    \item Our SAM3D-Part combines SAM3D-derived image prompts with input mesh features through an efficient pixel-aligned multimodal fusion paradigm. It further incorporates dense correspondence-based alignment and a part-cache mechanism to recover accurate part placement and improve consistency across sequential multi-part queries.

    \item Extensive experiments and ablations demonstrate that SAM3D-Part achieves better part quality, source alignment, and sequential consistency than existing methods.
\end{itemize}

%% file: sections/2_relatedwork.tex
\section{Related Work}
\label{sec:related_work}

\subsection{Object-level 3D Generation}
DreamFusion~\cite{poole2022dreamfusion} pioneers text-to-3D generation by optimizing a 3D representation with 2D diffusion priors via score distillation sampling. Subsequent methods improve optimization efficiency, geometry quality, and multi-view consistency~\cite{lin2023magic3d,chen2023fantasia3d,tang2023makeit3d,shi2023mvdream}, but still suffer from slow per-instance optimization and 3D inconsistency. Another line synthesizes multi-view images and reconstructs 3D assets from them~\cite{tang2024lgm,xu2024grm}, while single-pass feed-forward regressors enable faster image-to-3D generation at the cost of geometric fidelity~\cite{hong2024lrm,szymanowicz2024splatter}. Recently, native 3D generative models have been trained directly on 3D data with various representations~\cite{nichol2022point,jun2023shap}. VecSet-based latent diffusion~\cite{zhang20233dshape2vecset} has become a strong paradigm for scalable mesh generation, as shown by CLAY~\cite{zhang2024clay}, TripoSG~\cite{li2025triposg}, and Hunyuan3D-2.1~\cite{hunyuan3d2025hunyuan3d21}. Meanwhile, structured 3D latent models (e.g., TRELLIS~\cite{xiang2025structured}, Hi3DGen~\cite{ye2025hi3dgen}, and Sparc3D~\cite{li2026sparc3d}) further separate coarse structure prediction from high-resolution latent generation. Despite their strong ability to synthesize complete objects, these methods mainly focus on object-level generation, leaving selective extraction of user-specified components from existing meshes largely unexplored.

\subsection{Part-level 3D Generation}
Part-level generation aims to model objects as compositions of semantically or geometrically meaningful components. Some methods generate multiple parts jointly, often by predicting part layouts, bounding boxes, or structured latent variables before synthesizing part geometry~\cite{liu2024part123,chen2025partgen,yang2025omnipart,zhang2025bang,he2025unipart}. Other approaches formulate part generation as an automatic decomposition or reconstruction problem, where all parts of an object are produced in a single forward process or through sequential generation~\cite{lin2026partcrafter,tang2026partpacker,yan2024frankenstein,yang2025holopart,yan2025x,chen2026autopartgen}. These methods are valuable for part-aware object synthesis and structured shape modeling, but they usually assume that the goal is to recover or generate the full set of object parts. Image-conditioned methods (e.g., SAM3D~\cite{chen2025sam}) are closer to selective part generation, as they can focus on a visible region specified in an input view. However, because their conditions are mainly derived from rendered images, they may not fully preserve invisible geometry, volumetric occupancy, or the precise spatial relationship between the target part and the original mesh. In contrast, our work enables prompt-driven extraction of only the queried part from an existing mesh, producing a complete component aligned with the given object shape.

\subsection{3D Part Segmentation}
3D part segmentation has long been studied as a fundamental problem for understanding object structure. Early learning-based methods operate on point clouds, typically requiring dense part annotations for supervised training~\cite{qi2017pointnet,qi2017pointnet++,zhao2021point}. Although effective in specific domains, their generalization is limited by the scale and coverage of annotated 3D datasets. Recent methods leverage powerful 2D foundation models (e.g., SAM~\cite{kirillov2023sam}) to improve open-world 3D segmentation. A common strategy is to render a 3D object into multiple views, apply 2D segmentation or promptable segmentation in image space, and then lift the predicted masks back to the 3D surface~\cite{liu2023partslip,cen2023segment,abdelreheem2023satr,yang2023sam3d,liu2025partfield,yang2024sampart3d}. Meanwhile, promptable 3D segmentation methods extend the SAM paradigm to 3D, using user prompts such as points or boxes as 3D anchors to segment target parts~\cite{ma2025p3,li2026segvigen}. These methods greatly enhance flexibility and reduce the need for category-specific annotations, but their outputs are typically surface labels or masks rather than complete part meshes. As a result, they cannot directly provide complete components suitable for downstream editing, simulation, fabrication, or asset reuse. In contrast, our work uses SAM-derived part images as prompts to guide 3D part generation, achieving reusable and source-aligned part meshes.

%% file: sections/3_method.tex
\section{Methodology}
\label{sec:method}

\input{figures/pipeline}

\paragraph{Overview}
Given an object mesh $\mathcal{M}$ and a user-specified 2D part mask $m$ on one rendered view of $\mathcal{M}$, our goal is to generate the corresponding 3D part mesh $\mathcal{P}$. The generated part should be complete and aligned with the coordinate frame of the original mesh. We formulate this task as conditional 3D generation rather than surface segmentation. As demonstrated in \autoref{fig:pipeline}, our proposed SAM3D-Part takes both image-space prompts and mesh-space geometry as conditions, and directly generates the queried component as a standalone mesh. Built upon a two-stage sparse 3D generation framework, our method first predicts the coarse occupancy and source-frame correspondence of the target part, then refines it to achieve high-resolution geometry. This design allows the model to use simple user interactions while preserving the geometric context of the original mesh. A maintained cache volume of previously segmented parts enables consistent sequential part generation.

\subsection{Preliminaries}
\label{subsec:preliminaries}

\paragraph{Flow Matching}
Flow matching~\cite{liu2023flow} learns a continuous transformation from a simple source distribution to the data distribution. Let $x_0 \sim \mathcal{N}(0, I)$ be a Gaussian noise sample and $x_1$ be a data sample. In rectified flow, an intermediate sample at time $t \in [0,1]$ is defined as
\begin{equation}
    x_t = (1 - t)x_0 + t x_1 .
\end{equation}
The target velocity field is therefore
\begin{equation}
    v^{*} = x_1 - x_0 .
\end{equation}
A neural network $v_{\theta}(x_t, t, c)$ is trained to predict this velocity under condition $c$, using the objective
\begin{equation}
    \mathcal{L}_{\mathrm{FM}}
    =
    \mathbb{E}_{t, x_0, x_1, c}
    \left[
    \left\|
    v_{\theta}(x_t, t, c) - v^{*}
    \right\|_2^2
    \right].
\end{equation}
During inference, generation starts from $x_0 \sim \mathcal{N}(0, I)$ and follows the ordinary differential equation
\begin{equation}
    \frac{dx}{dt} = v_{\theta}(x_t, t, c)
\end{equation}
from $t=0$ to $t=1$, usually with Euler integration and classifier-free guidance.

\paragraph{TRELLIS}
TRELLIS~\cite{xiang2025structured} is a scalable two-stage framework for 3D generation based on structured 3D latents. Its first stage, called Sparse Structure (SS), predicts a coarse sparse occupancy structure. A flow-matching DiT (Diffusion Transformer~\cite{peebles2023scalable}) denoises latent tokens on a low-resolution 3D grid, and a pretrained occupancy VAE decodes them into a binary occupancy grid. This stage captures the main volumetric layout of the object. The second stage, called Structured Latent (SLat), performs generation only on occupied locations predicted by the first stage. It refines the geometry with high-resolution latent features and decodes them into the final mesh. This sparse-to-dense design reduces computation while preserving detailed geometry.

\paragraph{SAM3D}
SAM3D~\cite{chen2025sam} is closely related to our work and serves as the direct foundation of SAM3D-Part. It addresses the task of extracting a 3D object from a user-specified 2D mask in an input image. Following the two-stage generation pipeline of TRELLIS, it uses image-space conditions, including the rendered RGB image, the user mask, and a predicted point map, to guide 3D object generation. However, its conditioning primarily relies on 2D observations and does not explicitly use the input mesh geometry. In addition, it predicts a global 7-DoF pose with a single pose head to map the generated object back to the world coordinate frame. In contrast, our task requires extracting a specific part from an existing mesh, where the mesh geometry provides important constraints. Therefore, we extend this framework with mesh-aware conditioning, dense correspondence prediction, and an iterative part-cache mechanism.

\subsection{Multi-modal Conditioning for Part-aware Generation}
\label{subsec:multimodal_conditioning}

To obtain a specified part mesh, our SAM3D-Part uses both 2D user prompts and 3D mesh geometry as conditions. The 2D prompt presents the user intent and part information, while the 3D mesh condition provides the full object geometry, including regions that may be invisible from the selected view. We encode these inputs into two condition streams: a 2D condition stream from rendering and user masks, and a 3D condition stream from the source mesh and the part cache. The two streams are concatenated along the token dimension and injected into the DiT through cross-attention.

\paragraph{View Rendering and 2D Prompt Encoding}
We first render the input mesh $\mathcal{M}$ from a user-selected view. The rendering produces an RGB image, a foreground mask, and a point map. The point map records the 3D coordinate corresponding to each foreground pixel, and sets background pixels to a constant value (i.e., $-1$). The user then specifies the target part on the rendered image through simple interactions such as clicks, and a SAM-based segmentor produces the 2D part mask~\cite{yang2023sam3d,ravi2025sam}.

To capture both local part details and global object context, we encode two views for each 2D signal: a cropped view around the part mask and a full rendered view. The RGB image and the mask are encoded by a frozen DINOv2 image encoder~\cite{oquab2024dinov2}, while the point map is encoded by a lightweight trainable patch encoder. Instead of concatenating the three modalities as separate token sequences, we fuse them at the same spatial patch locations. In this way, each token contains appearance, user attention, and 3D position information for the same image region. This cross-modal fusion keeps the token size compact and makes the 2D condition easier for the DiT to use.

\paragraph{Global Mesh Encoding}
The 2D prompt alone is not sufficient for part extraction, because it only observes the object from one view. To provide complete 3D information, we encode the input mesh $\mathcal{M}$ with a pretrained Hunyuan3D-2.1 ShapeVAE encoder~\cite{hunyuan3d2025hunyuan3d21}. Specifically, we sample $8,1920$ surface points from $\mathcal{M}$, including both uniformly sampled points and curvature-weighted points, and feed them into the frozen encoder with corresponding normal and sharp-edge labels. The encoder produces a set of 3D anchors with latent features. Each anchor feature is concatenated with its positional encoding, giving a compact token representation of the input mesh. These mesh tokens allow the generation model to directly attend to the geometry of the original object, instead of reconstructing it only from the rendered view.

\paragraph{Part-Cache Encoding}
For the iterative part extraction, we maintain a binary voxel cache $\mathcal{C}_i$ for each extraction step $i$ to record the parts extracted in previous steps, i.e.,
\begin{equation}
    \mathcal{C}_i =
    \mathrm{OR}\left(
    \mathrm{Voxelize}(\mathcal{P}_1),
    \ldots,
    \mathrm{Voxelize}(\mathcal{P}_{i-1})
    \right).
\end{equation}
This cache tells the model which regions have already been extracted. Instead of training a separate cache encoder, we attach the cache information to the global mesh tokens. For each mesh anchor, we query the corresponding voxel in $\mathcal{C}_i$ via nearest-voxel lookup and obtain a binary flag indicating whether this location belongs to a previously extracted part. This flag is concatenated with the anchor latent and position feature, and then projected to the DiT condition dimension. As a result, each 3D condition token contains local geometry, spatial position, and extraction history. 
This design naturally supports iterative sub-part extraction.

\subsection{Two-stage Part Generation and Global Alignment}
\label{subsec:generation_alignment}

Given the multi-modal conditions, SAM3D-Part follows a two-stage generation process, as shown in Fig.~\ref{fig:pipeline}. The first stage predicts the sparse structure of the target part and its dense correspondence to the source mesh. The second stage refines the part geometry on the predicted structure and decodes the final mesh. Both stages are trained with the flow matching objective introduced in Sec.~\ref{subsec:preliminaries}.

\paragraph{Stage 1: Sparse Structure and Dense Correspondence Generation}
The first stage generates the coarse volumetric structure of the queried part. A flow-matching DiT denoises a 3D latent grid under the conditions from the 2D prompt, point map, source mesh tokens, and part-cache tokens. The output latent is decoded by two heads. The first head predicts a binary occupancy grid, which describes the spatial support of the target part. The second head predicts a dense XYZ volume, where each occupied voxel is assigned a corresponding 3D coordinate in the source mesh frame.

This dense XYZ prediction is important for robust alignment. A direct pose regression head would provide only a small number of supervision signals for each training sample, making pose learning unstable. Instead, our model predicts source-frame coordinates for all occupied voxels, providing dense voxel-level supervision. Since each occupied voxel has both a local coordinate and a predicted source-frame coordinate, their correspondence is already known and no nearest-neighbor matching is required.

\paragraph{Closed-Form Source-Frame Alignment}
The generated part is represented in a normalized local coordinate system. During training, each ground-truth part is transformed from the source frame to this local frame using only centering and isotropic scaling, without rotation canonicalization. Thus, its local axes inherit the source-mesh axes, while the dense XYZ branch is supervised in the source frame. Accordingly, we estimate only an isotropic scale and a translation; this orientation consistency is learned rather than guaranteed at inference.
Let $p_{\mathrm{local}}(v)$ be the local coordinate of an occupied voxel $v$, and let $p_{\mathrm{global}}(v)$ be the source-frame coordinate predicted by the dense XYZ head. We solve
\begin{equation}
    \min_{s,o}
    \sum_{v}
    \left\|
    s\cdot p_{\mathrm{local}}(v) + o - p_{\mathrm{global}}(v)
    \right\|_2^2 ,
\end{equation}
where $s$ is the isotropic scale and $o$ is the translation. This least-squares problem has a closed-form solution. Let $\bar{p}_{\mathrm{local}}$ and $\bar{p}_{\mathrm{global}}$ be the centroids of the two point sets. The translation is
\begin{equation}
    o = \bar{p}_{\mathrm{global}} - s\cdot \bar{p}_{\mathrm{local}},
\end{equation}
and the scale is obtained by projecting the centered local coordinates onto the centered global coordinates. The recovered $(s,o)$ is then used to place the generated part back into the coordinate frame of the source mesh. This alignment is not an ICP procedure~\cite{chetverikov2002trimmed}, because the correspondences are directly predicted by the model.

\paragraph{Stage 2: Structured Latent Refinement}
After the sparse structure and alignment are determined, the second stage refines the detailed geometry of the part. It performs structured latent denoising only on the occupied voxels predicted by Stage 1. Compared with Stage 1, this stage focuses on local geometric details rather than deciding which region should be extracted. We also inject the same global mesh condition into the Stage-2 DiT, so that the refined geometry remains consistent with the surface and shape of the source object. Finally, the structured latent is decoded into a complete part mesh with geometry and appearance.

\paragraph{Training Objective}
Both stages are trained with the same flow matching objective:
\begin{equation}
    \mathcal{L}_{\mathrm{FM}}
    =
    \mathbb{E}_{t, x_0, x_1, c}
    \left[
    \left\|
    v_{\theta}(x_t, t, c) - (x_1 - x_0)
    \right\|_2^2
    \right],
\end{equation}
where $x_1$ is the ground-truth latent target, $x_0 \sim \mathcal{N}(0,I)$ is Gaussian noise, and $c$ denotes the multi-modal conditions. In Stage 1, $x_1$ contains the latent targets for both occupancy and dense XYZ prediction. In Stage 2, $x_1$ is the structured latent of the target part. During inference, we sample from Gaussian noise and integrate the learned velocity field with classifier-free guidance to obtain the final part mesh.

%% file: figures/pipeline.tex
\begin{figure*}
  \centering
  \includegraphics[width=1.\linewidth]{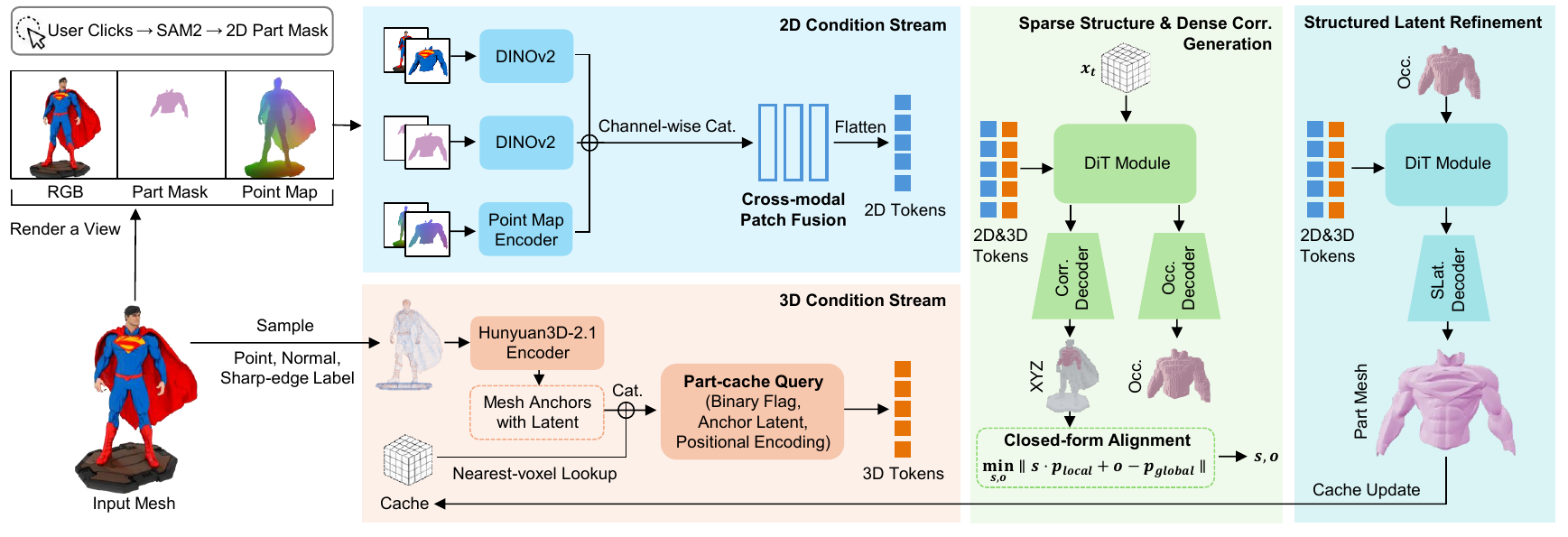}
  \caption{SAM3D-Part pipeline. Given an input mesh and user-specified image-space prompts, SAM3D-Part encodes the rendered RGB image, part mask, point map, and source mesh geometry into 2D and 3D condition tokens. These tokens guide a two-stage generator that first predicts the coarse occupancy and dense source-frame correspondences of the selected part, and then refines them into a high-resolution part mesh. A cache volume storing previously generated parts is further used to support consistent sequential part generation.}
  \label{fig:pipeline}
\end{figure*}

%% file: sections/4_experiment.tex
\section{Experiments}
\label{sec:experiments}

\input{tables/quant_compare}
\input{tables/partobjaverse_tiny}
\input{tables/ablation}
\input{tables/viewpoint_click_ablation}
\input{tables/efficiency}
\input{figures/ablation}

\subsection{Experimental Setup}
\label{sec:setup}

\paragraph{Implementation Details}
Our model uses a two-stage design. Stage 1 is a sparse-structure DiT with $24$ transformer blocks. It predicts $16^{3}$ occupancy grids and xyz latent grids. Stage 2 is an SLat DiT that converts the Stage 1 occupancy output into a high-resolution mesh, conditioned on the input image, mask, and global mesh. Specifically, for conditioning, the 2D branches share a frozen DINOv2 ViT-L backbone with $518/14$ patches. The xyz point-map branch uses a $296/8$ PointPatchEmbed. The 3D branch encodes the input mesh with the Hunyuan3D-2.1 ShapeVAE, which samples $81{,}920$ surface points, including $40{,}960$ uniformly sampled points and $40{,}960$ curvature-weighted points, and produces $4{,}096$ mesh anchor tokens.
We train Stage 1 for $60$k iterations with a global batch size of $128$ on $8$ H100 GPUs, using AdamW with a peak learning rate of $5{\times}10^{-5}$. Stage 2 is fine-tuned for $40$k iterations with the same optimizer. During inference, we use $25$ sampling steps for Stage 1 and $12$ sampling steps for Stage 2 on a single H100 GPU in fp16.

\paragraph{XYZ VAE}
The XYZ latent predicted by Stage 1 is decoded into a $64^{3}$ per-voxel correspondence map using a sparse-structure XYZ VAE, adapted from the TRELLIS Stage-1 occupancy VAE~\cite{xiang2025structured}. We retain the encoder-decoder backbone and the $16^{3}$ latent grid with $8$ channels, then make three modifications: (i) We replace the $1$-channel occupancy input with $4$-channel XYZ coordinates, enabling the VAE to encode per-voxel correspondences instead of binary occupancy; (ii) To help the decoder focus on occupied voxels, we make it \emph{occupancy-conditioned}. Specifically, at the decoder bottleneck, we concatenate the $8$-channel occupancy latent from the frozen SAM3D occupancy VAE with the $8$-channel XYZ latent along the channel dimension; (iii) We simplify the output head to a shared $\mathrm{Norm}{+}\mathrm{SiLU}$ block followed by a $3$-channel $3{\times}3{\times}3$ convolution and a $[-1,1]$ clamp. We remove the additional non-linear layers in the original head because our target is a continuous coordinate rather than a logit.

We train the XYZ VAE with an $L_1$ reconstruction loss on XYZ values, evaluated only at occupied voxels, together with a TV smoothness regularizer weighted by $0.01$ and a KL term weighted by $10^{-3}$. To prevent large parts from dominating the loss, we divide the per-sample weight by the bounding-box diagonal of each part. This size re-weighting gives small parts a comparable training signal. The XYZ VAE is trained on the same dataset as the DiT for $70$k iterations using AdamW with a learning rate of $1{\times}10^{-4}$, a global batch size of $128$, and an EMA decay of $0.9999$. We use the EMA weights at inference time.

\paragraph{Datasets}
We curate part-level training data from five public 3D asset collections, i.e., Objaverse~\cite{deitke2023objaverse}, Objaverse-XL~\cite{deitke2023objaversexl}, ABO~\cite{collins2022abo}, 3D-FUTURE~\cite{fu20213d}, and HSSD~\cite{khanna2024habitat}. Each GLB file often contains a native part structure, which we use as the basis for data curation. We first discard assets with fewer than $2$ parts or more than $32$ parts, since they are either not meaningfully decomposed or are often over-segmented and scene-like. This gives an initial pool of about $446{\rm k}$ assets. We then split each sub-mesh by connectivity and merge small fragments into nearby larger parts based on collision proximity, leading to more balanced part decompositions. Finally, annotators manually remove assets with unreasonable decompositions, such as parts crossing semantic boundaries, duplicated parts, or visually indistinguishable parts, resulting in $187{\rm k}$ verified assets.

For evaluation, we use three sets from different sources. The first is a \emph{multi-part test set} seperate from training set containing $500$ assets with ground-truth parts. The second is the \emph{PartObjaverse-Tiny}, a public dataset mostly used by 3D part segmentation and part-level generation. The third is an \emph{in-the-wild set} generated from commercial 3D generator. Since it does not provide ground truth, it is not used for metric computation.


\paragraph{Baselines}
We compare against four representative baselines, i.e., SAM3D~\cite{chen2025sam}, the two-stage pipelines HoloPart~+~P3-SAM~\cite{yang2025holopart} and X-Part~+~P3-SAM~\cite{yan2025x}, and the end-to-end method OmniPart~\cite{yang2025omnipart}. The HoloPart/X-Part pipelines first use P3-SAM to segment the input mesh in 3D and then complete each segment into a part mesh, while SAM3D and OmniPart directly predict part meshes from the 2D condition. 
For a fair comparison, all methods are queried using the same 2D target mask. OmniPart receives the mask directly, while foreground points sampled from the mask are lifted to the source mesh as 3D prompts for P3-SAM. Since the P3-SAM-based methods output a full decomposition, we select the predicted component with the highest GT-part BBox IoU. This oracle selection gives the decomposition baselines their most favorable match for evaluation. All baselines use publicly released checkpoints and inputs generated from the same source assets.

\paragraph{Evaluation Metrics}
We evaluate part geometry with four metrics: bounding-box IoU (\emph{BBox IoU}), Chamfer Distance (\emph{CD}), and F1-scores at $\tau{=}0.05$ and $\tau{=}0.1$. Before evaluation, we normalize all predicted and ground-truth parts to the ground-truth bounding box, which is scaled to the $[-0.5,0.5]^{3}$ cube. This removes the effects of global scale and canonicalization differences. We also evaluate Stage 1 efficiency using the number of cross-attention condition tokens $N_{c}$, cross-attention FLOPs, latency, and peak VRAM usage.

\subsection{Comparison with State-of-the-Art}
\label{sec:compare}
Tab.~\ref{tab:quant_compare_main} reports the quantitative results on the multi-part test set. Our method achieves the best performance on all metrics: BBox IoU increases to $\mathbf{0.641}$, compared with the next-best score of $0.552$ from OmniPart; CD decreases to $\mathbf{0.053}$, compared with $0.192$ from OmniPart; and F1-$0.1$ improves from $0.756$ to $\mathbf{0.849}$. This trend is consistent under both strict and relaxed F1 thresholds, showing that the improvement is not limited to a specific tolerance level but reflects a broader gain in geometric accuracy. 
Tab.~\ref{tab:partobjaverse_tiny} reports the quantitative results on PartObjaverse-Tiny dataset. Our method still performs strongly on this independent benchmark, supporting that the method generalizes beyond our curated benchmark.

As shown in Fig.~\ref{fig:comparison_dataset}, visual results also demonstrate the superiority of our method. The baselines exhibit three common failure cases: (i) The two-stage \emph{segmentation~+~completion} pipelines (P3-SAM~+~X-Part and P3-SAM~+~HoloPart) inherit the segmentation behavior of P3-SAM. As a result, their segmentations are hard to control and often overly
fragmented, and the completion stage further amplifies these errors. (ii) Although OmniPart achieves competitive BBox IoU, it tends to produce overly smooth part geometry and misses fine details from the input mesh. (iii) SAM3D does not use a 3D condition, so its predictions often drift away from the input mesh and recover generic parts that do not remain on the original surface. In contrast, our results stay well aligned with the input mesh while preserving fine-grained shape details, which explains the large improvements in CD and F1.

Fig.~\ref{fig:comparison_inthewild} and Fig.~\ref{fig:comparison_more_case} further evaluate in-the-wild inputs that are outside the training distribution. Fig.~\ref{fig:comparison_inthewild} uses the same six-method layout and shows that our method remains robust on stylized characters, cartoon weapons, and furniture. The baselines either miss entire parts or collapse them into coarse shapes, whereas our method preserves both the part decomposition and the surface alignment. These results suggest that our method can generalize to animals, humans, mechanical parts, plants, and buildings. 

\subsection{Ablation Study}
\label{sec:ablation}

We study our method from two aspects: the contribution of each component to accuracy (Tab.~\ref{tab:ablation}) and the efficiency benefit of cross-modal token fusion (Tab.~\ref{tab:efficiency}). All ablation studies are conducted on the multi-part test set, using the same training and inference settings as in the main experiment.

\subsubsection{Component Ablation}
\label{sec:ablation_acc}

Tab.~\ref{tab:ablation} shows a cumulative ablation. We start from a finetuned SAM3D baseline and add our design choices one at a time. Fig.~\ref{fig:ablation} shows the corresponding visual results for two assets, where each row matches rows (a)-(f) in Tab.~\ref{tab:ablation}.

\textit{(a) Finetuned SAM3D.}
We first finetune SAM3D on our training dataset. This model uses only the 2D RGB image and mask as input. It benefits from in-domain training, achieving a BBox IoU of $0.275$ and a CD of $0.269$. However, without any 3D condition, its predictions can drift away from the input surface, as shown by the rocket and bed examples in Fig.~\ref{fig:ablation}.

\textit{(b) +Mesh conditioning.}
We then encode the input mesh with ShapeVAE anchors and inject them into cross-attention. This gives the largest improvement in Tab.~\ref{tab:ablation}: BBox IoU increases from $0.275$ to $0.426$ ($+0.151$), while CD decreases from $0.269$ to $0.134$. The rocket's top and base are better separated, and the bed begins to show distinct pillow and blanket regions. This shows that 3D anchors provide useful surface information that 2D conditions alone cannot capture.

\textit{(c) +Token fusion.}
The 2D inputs, including RGB, mask, and pointmap, are encoded separately across two views. Since directly using all tokens makes cross-attention costly and less focused, we fuse tokens from the three 2D modalities at the same patch location and view. This further improves BBox IoU to $0.518$ and CD to $0.128$. As shown in Tab.~\ref{tab:efficiency}, it also reduces the Stage 1 cost below the SAM3D baseline. Visually, the rocket-body boundary becomes
cleaner, and the bed surface shows more accurate pillow folds.

\textit{(d) +XYZ branch.}
We replace the single-vector pose head with an \emph{XYZ branch}. In this design, Stage 1 predicts a per-voxel XYZ correspondence latent from the generated part to the source space. The part scale and translation are then computed from these dense voxel-level votes, rather than regressed directly. This design is more robust under partial visibility. It improves the BBox IoU to $0.611$ and CD to $0.067$, with clearer placement of the rocket body and base and better contact between the blanket and mattress in Fig.~\ref{fig:ablation}.

\textit{(e) +Part cache}
For sequential generation, we voxelize previously generated parts into a binary cache and query the cache at each mesh anchor. The resulting binary flag is appended to the anchor feature, providing later queries with the extraction history. This improves BBox IoU to $0.625$ and CD to $0.061$ and reduces conflicts between generated parts.

\textit{(f) +2-stage with mesh condition.}
Finally, we also provide the input mesh as a 3D condition to Stage 2 during SLat refinement. This gives the full model, with a BBox IoU of
$\mathbf{0.641}$, CD of $\mathbf{0.053}$, and F1-$0.1$ of $\mathbf{0.849}$. The Stage 2 mesh condition mainly improves fine surface details, such as the sharper rocket base and clearer blanket folds in the last column of Fig.~\ref{fig:ablation}.


\subsubsection{Viewpoint and Click Robustness Ablation}
\label{sec:ablation_viewpoint_click}
Our training protocol already exposes the model to some variations: rendering views are randomly sampled, and the input masks are SAM2-predicted masks from GT-derived prompts rather than GT masks. We keep masks with 2D IoU > 0.6 during training to remove invalid or highly ambiguous prompts. To verify this robustness, we conducted a small-scale study by sampling 100 cases from the test set. For viewpoint robustness, we sample 6 valid views per target part, where the target part is visible, use the gt mask for each view, and evaluate the generated 3D part. For click robustness, we fix the view and sample 6 different click seeds on the visible target part, again using SAM2 masks. As shown in Tab.~\ref{tab:viewpoint_click_ablation}, the results show stable performance across views and clicks.

\subsubsection{Token Fusion Ablation}
\label{sec:ablation_eff}

Token fusion is not only useful for accuracy but also important for efficiency, as shown in Tab.~\ref{tab:efficiency}. We measure the Stage 1 cross-attention cost on a single H100 in fp16, with batch size $1$, $25$ sampling steps, and classifier-free guidance. We report the mean over $10$ timed runs after $2$ warm-up runs. Tab.~\ref{tab:efficiency} shows three main results: (i) The SAM3D baseline attends to $N_{c}{=}8{,}218$ tokens and uses $4.13$\,T cross-attention FLOPs, with a latency of $11{,}246.0$~ms and a peak VRAM of $22.09$~GB. (ii) Directly adding our $4{,}096$ mesh anchor tokens to the 2D conditions increases $N_{c}$ to $12{,}314$ ($+50\%$), FLOPs to $5.99$\,T ($+45\%$), latency to $13{,}416.1$~ms ($+19\%$), and peak VRAM to $23.73$~GB ($+7.4\%$). Thus, the 3D condition would add a clear efficiency cost without using token fusion. (iii) Cross-modal patch fusion merges the three 2D modalities at the same patch locations into one shared token set. This reduces $N_{c}$ to $6{,}836$, which is lower than the SAM3D baseline even with the added 3D modality. As a result, our full model uses $3.51$\,T FLOPs ($-15\%$ vs.\ SAM3D), $10{,}450.7$~ms latency ($-7\%$), and $21.53$~GB peak VRAM ($-2.5\%$). Together with the accuracy gain in Tab.~\ref{tab:ablation}, this shows that token fusion provides a double benefit: it removes redundant 2D tokens while covering the extra cost of the $4{,}096$ mesh anchor tokens.

\subsection{Limitations and Failure Cases}
\label{sec:limit}
While our SAM3D-Part can generate high-quality specified 3D parts, it still struggles with extremely complex cases. Fig.~\ref{fig:failure} shows a failure case: a sculpted figure holding a hanging lantern and wrapped by thin vines. This example contains \emph{thin lattice-like structures}, such as vine stems, leaf veins, and the open lantern frame, as well as \emph{strongly overlapping parts}, where vines wind around both the figure and the lantern. In this setting, structures thinner than a voxel cannot be captured by the $64^{3}$ sparse representation or the Hunyuan3D ShapeVAE bottleneck. As a result, the lantern frame and the smallest vine branches are smoothed out or merged with nearby surfaces. The overlapping geometry also makes the queried 2D mask ambiguous across different depth layers. The model may therefore assign geometry to the wrong layer or split one part into disconnected fragments. Addressing these challenges may require higher-resolution sparse representations, multi-view mask supervision, and explicit modeling of part-to-part contact, which we leave for future work.


%% file: tables/quant_compare.tex
\begin{table}[htb]
\centering
\caption{Quantitative comparison of part generation. CD and F1-$\tau$ are computed in GT-bbox-normalized space. \textbf{Bold}: the best result for each metric.}
\label{tab:quant_compare_main}
\setlength{\tabcolsep}{4pt}
\small
\begin{tabular}{l cccc}
\toprule
Method & BBox IoU $\uparrow$ & CD $\downarrow$ & F1-0.05 $\uparrow$ & F1-0.1 $\uparrow$ \\
\midrule
SAM-3D-Objects (Meta)        & 0.225 & 0.375 & 0.221 & 0.388 \\
HoloPart $+$ P3-SAM          & 0.291 & 0.452  & 0.445 & 0.521 \\
X-Part $+$ P3-SAM            & 0.458 & 0.282  & 0.608 & 0.674 \\
OmniPart                     & 0.552 & 0.192  & 0.666 & 0.756 \\
\textbf{Ours (full)}         & \textbf{0.641}	& \textbf{0.053} & \textbf{0.716}	& \textbf{0.849} \\
\bottomrule
\end{tabular}
\end{table}

%% file: tables/partobjaverse_tiny.tex
\begin{table}[htb]
\centering
\caption{
        Quantitative comparison on PartObjaverse-Tiny. \textbf{Bold}: the best result for each metric.
        }
\label{tab:partobjaverse_tiny}
\setlength{\tabcolsep}{4pt}
\small
\begin{tabular}{l cccc}
\toprule
Method & BBox IoU $\uparrow$ & CD $\downarrow$ & F1-0.05 $\uparrow$ & F1-0.1 $\uparrow$ \\
\midrule
SAM-3D-Objects (Meta) & 0.255 & 0.294 & 0.240 & 0.392 \\
HoloPart $+$ P3-SAM   & 0.270 & 0.114 & 0.531 & 0.617 \\
X-Part $+$ P3-SAM     & 0.317 & 0.111 & 0.547 & 0.634 \\
OmniPart              & 0.473 & 0.065 & 0.585 & 0.782 \\
\textbf{Ours (full)}  & \textbf{0.688} & \textbf{0.054} & \textbf{0.694} & \textbf{0.851} \\
\bottomrule
\end{tabular}
\end{table}

%% file: tables/ablation.tex
\begin{table}[htb]
\centering
\caption{Ablation on the multi-part test set. CD and F1-$\tau$ are computed in the GT-bbox-normalized space. \textbf{Bold}: the best result for each metric.}
\label{tab:ablation}
\setlength{\tabcolsep}{4pt}
\small
\begin{tabular}{l cccc}
\toprule
Configuration & BBox $\uparrow$ & CD $\downarrow$ & F1-0.05 $\uparrow$ & F1-0.1 $\uparrow$ \\
\midrule
(a) Finetuned SAM3D      & 0.275 & 0.269 & 0.311 & 0.453 \\
(b) \;\;$+$ Mesh conditioning           & 0.426 & 0.134 & 0.510 & 0.664 \\
(c) \;\;$+$ Token fusion           & 0.518 & 0.128 & 0.592 & 0.731 \\
(d) \;\;$+$ XYZ branch & 0.611 & 0.067 & 0.697 & 0.819  \\
(e) \;\;$+$ Part cache         & 0.625 & 0.061  & 0.711 & 0.839\\
(f) \;\;$+$ 2-stage w. mesh condition    & \textbf{0.641}	& \textbf{0.053} & \textbf{0.716}	& \textbf{0.849} \\
\bottomrule
\end{tabular}
\end{table}

%% file: tables/viewpoint_click_ablation.tex
\begin{table}[t]
\centering
\caption{
        Ablation on the robustness to viewpoint and click variations.
        }
\label{tab:viewpoint_click_ablation}
\setlength{\tabcolsep}{4pt}
\small
\begin{tabular}{l cccc}
\toprule
Study & BBox IoU $\uparrow$ & CD $\downarrow$ & F1-0.05 $\uparrow$ & F1-0.1 $\uparrow$ \\
\midrule
Viewpoint & $0.616 \pm 0.098$ & $0.057 \pm 0.020$ & $0.708 \pm 0.097$ & $0.838 \pm 0.075$ \\
Click     & $0.609 \pm 0.065$ & $0.067 \pm 0.024$ & $0.686 \pm 0.053$ & $0.819 \pm 0.051$ \\
\bottomrule
\end{tabular}
\end{table}

%% file: tables/efficiency.tex

\begin{table}[t]
\centering
\caption{
Effect of token fusion on Stage-1 cross-attention efficiency.
Our fusion reduces token count and computational cost while incorporating additional 3D conditions.
}
\label{tab:efficiency}
\setlength{\tabcolsep}{4pt}
\small
\renewcommand{\arraystretch}{1.2}
\begin{tabular}{lccc}
\toprule
\textbf{Metric} & \textbf{SAM3D} & \textbf{Ours} & \textbf{Ours} \\
                &                & \textbf{(w/o fusion)}   & \textbf{(w/ fusion)} \\
\midrule
Tokens $N_c$        & 8\,218  & 12\,314 & \textbf{6\,836} \\
Cross-attn FLOPs    & 4.13\,T & 5.99\,T & \textbf{3.51\,T} \\
Latency (ms)        & 11\,246.0 & 13\,416.1 & \textbf{10\,450.7} \\
Peak VRAM (GB)      & 22.09   & 23.73   & \textbf{21.53} \\
\bottomrule
\end{tabular}
\end{table}

%% file: figures/ablation.tex
\begin{figure}
\centering
  \includegraphics[trim=0cm 0cm 0cm 0cm, clip=true,width=0.99\linewidth]{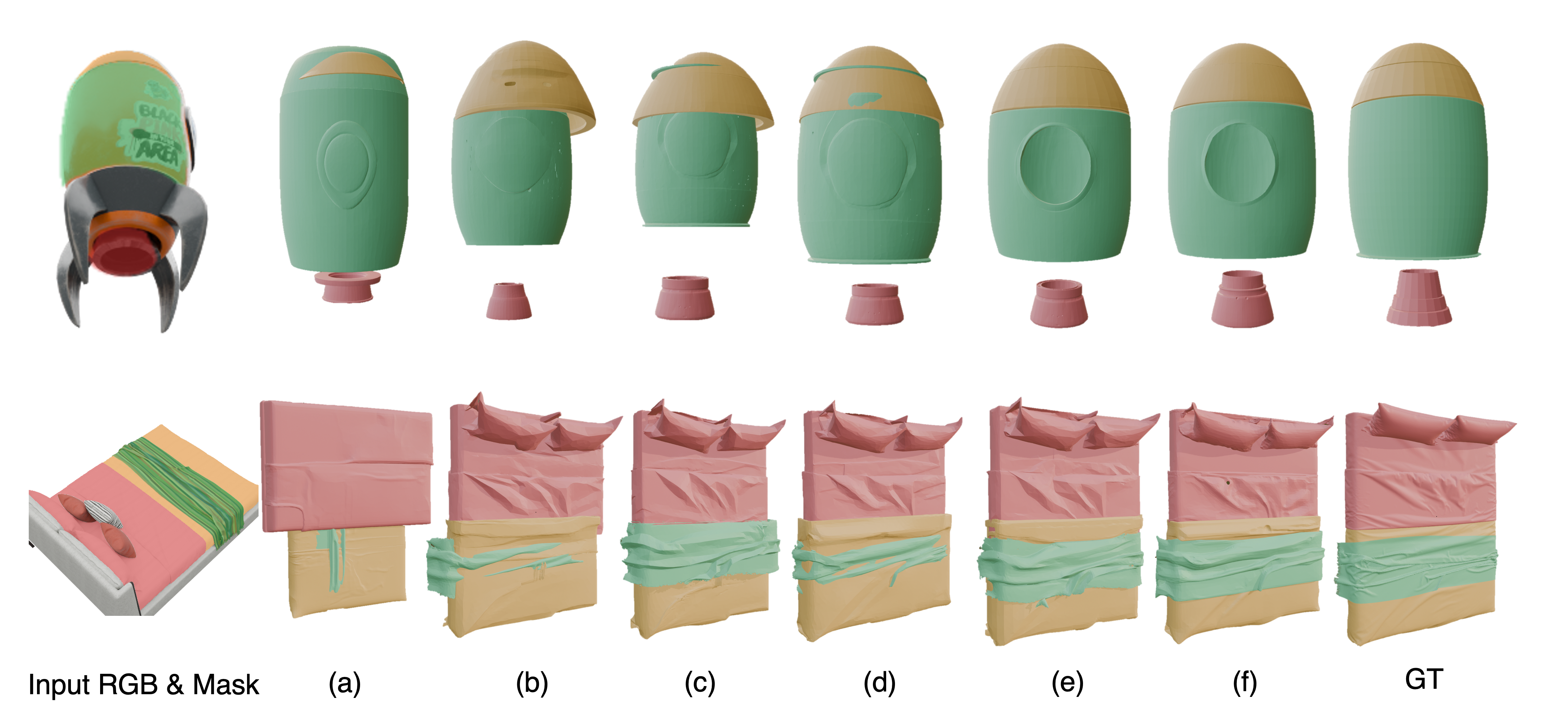}
  \caption{Visual results of ablation studies for the design of SAM3D-Part.}
  \label{fig:ablation}
\end{figure}

%% file: sections/5_conclusion.tex
\section{Conclusion}
\label{sec:conclusion}
We presented SAM3D-Part, a prompt-driven framework for generating a specified
3D part from a source mesh. Given a user prompt, SAM3D-Part returns the queried
part as a complete mesh aligned with the source. It combines SAM-based 2D part
prompts with the input mesh through pixel-aligned multimodal fusion, and places
the generated part back into the source frame using dense voxel-level XYZ
correspondences. A part cache further improves consistency across sequential
multi-part queries.
Experiments show that SAM3D-Part achieves strong part quality, source
alignment, and sequential consistency. The ablation study also shows that
cross-modal token fusion improves both accuracy and efficiency, reducing the
Stage~1 cost below the SAM-3D-Objects baseline.
SAM3D-Part still struggles with thin lattice-like structures and heavily
overlapping or nested parts. We plan to address these limitations with
higher-resolution sparse representations, multi-view mask supervision, and
explicit part-to-part contact modeling. We hope this prompt-driven and
source-aligned design can support future part-level workflows in 3D editing,
animation, and fabrication.

%% file: sections/6_figs_only.tex
\input{figures/quali_compare}
\input{figures/quali_compare_in_the_wild}
\input{figures/failure_case}

\begin{figure*}
\centering
  \includegraphics[trim=0cm 0cm 0cm 0cm, clip=true, width=0.78\linewidth]{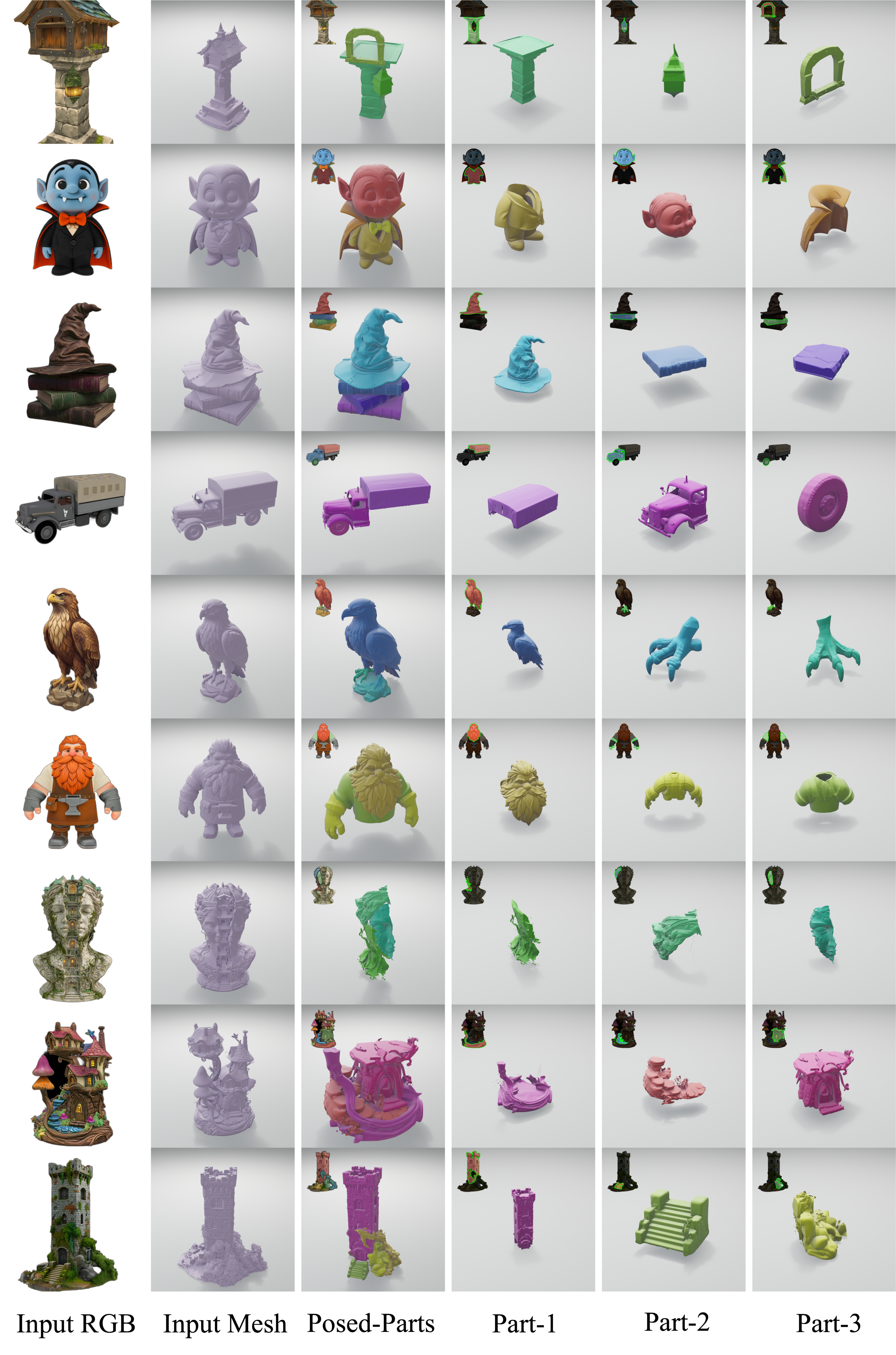}
  \caption{More in-the-wild results of our method. For each example, we show the input RGB image, input mesh, assembled posed parts, and three generated parts laid out separately.
}

  \label{fig:comparison_more_case}
\end{figure*}

%% file: figures/quali_compare.tex
\begin{figure*}[htb]
\centering
  \includegraphics[trim=0cm 0cm 0cm 0cm, clip=true,width=0.9\linewidth]{figures/sam3dpart-dataset-compressed.pdf}
  \caption{Qualitative comparison on the multi-part test set. For each example
(row), we show the input RGB and mask (left), the part predictions of all
baselines, our method, and the GT (right). For each method, we visualize both the parts placed in the source object and the parts laid out separately.
}
  \label{fig:comparison_dataset}
\end{figure*}

%% file: figures/quali_compare_in_the_wild.tex
\begin{figure*}
\centering
  \includegraphics[trim=0cm 0cm 0cm 0cm, clip=true,width=0.9\linewidth]{figures/sam3dpart-in-the-wild-compressed.pdf}
  \caption{Qualitative comparison on in-the-wild assets (no GT available).
For each example (row), we show the input RGB and mask (left), the source
mesh, and the part predictions of all baselines and our method. For each method, we visualize both the parts placed in the source object and the parts laid out separately.
}
  \label{fig:comparison_inthewild}
\end{figure*}

%% file: figures/failure_case.tex
\begin{figure*}[htb]
\centering
  \includegraphics[trim=0cm 0cm 0cm 0cm, clip=true,width=0.65\linewidth]{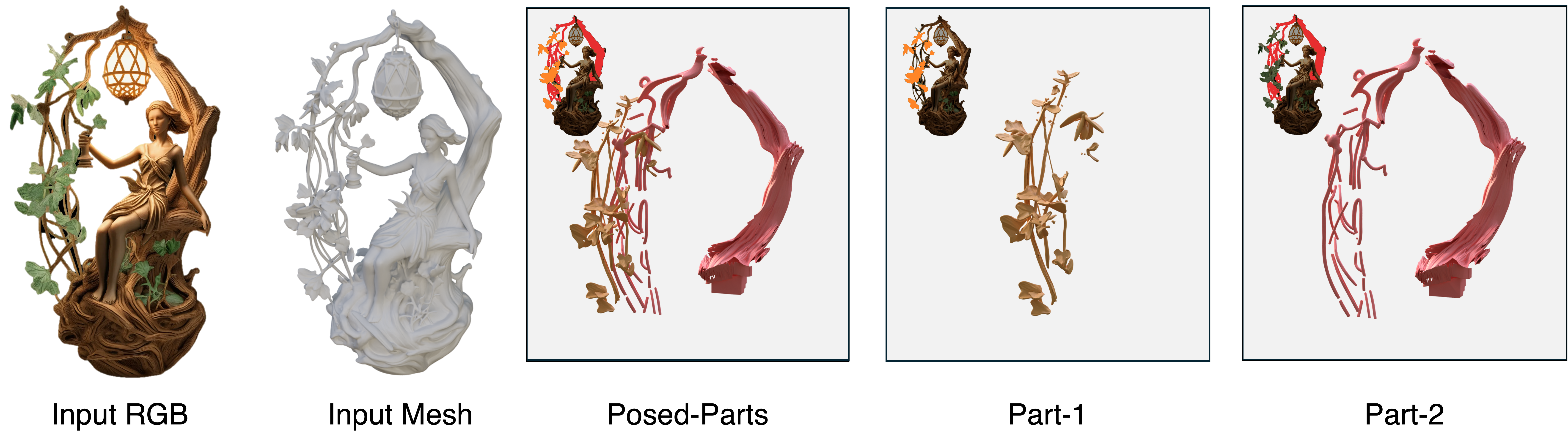}
  \caption{Failure case. From left to right: input RGB, input mesh, and the two parts predicted by our method. Our method struggles with assets containing thin, lattice-like structures (e.g., vines, leaf veins, and openwork lantern frames), where parts are heavily entangled.
}
  \label{fig:failure}
\end{figure*}

%% file: reference.bib
@String{Computer = "{IEEE} Computer" }

@String{Springer = "Springer-Verlag" }

@article{poole2022dreamfusion,
  title={Dreamfusion: Text-to-3d using 2d diffusion},
  author={Poole, Ben and Jain, Ajay and Barron, Jonathan T and Mildenhall, Ben},
  journal={arXiv preprint arXiv:2209.14988},
  year={2022}
}

@inproceedings{lin2023magic3d,
  title={Magic3d: High-resolution text-to-3d content creation},
  author={Lin, Chen-Hsuan and Gao, Jun and Tang, Luming and Takikawa, Towaki and Zeng, Xiaohui and Huang, Xun and Kreis, Karsten and Fidler, Sanja and Liu, Ming-Yu and Lin, Tsung-Yi},
  booktitle={Proceedings of the IEEE/CVF conference on computer vision and pattern recognition},
  pages={300--309},
  year={2023}
}

@inproceedings{chen2023fantasia3d,
  title={Fantasia3d: Disentangling geometry and appearance for high-quality text-to-3d content creation},
  author={Chen, Rui and Chen, Yongwei and Jiao, Ningxin and Jia, Kui},
  booktitle={Proceedings of the IEEE/CVF international conference on computer vision},
  pages={22246--22256},
  year={2023}
}

@inproceedings{tang2023makeit3d,
  title={Make-it-3d: High-fidelity 3d creation from a single image with diffusion prior},
  author={Tang, Junshu and Wang, Tengfei and Zhang, Bo and Zhang, Ting and Yi, Ran and Ma, Lizhuang and Chen, Dong},
  booktitle={Proceedings of the IEEE/CVF international conference on computer vision},
  pages={22819--22829},
  year={2023}
}

@article{shi2023mvdream,
  title={Mvdream: Multi-view diffusion for 3d generation},
  author={Shi, Yichun and Wang, Peng and Ye, Jianglong and Long, Mai and Li, Kejie and Yang, Xiao},
  journal={arXiv preprint arXiv:2308.16512},
  year={2023}
}

@article{nichol2022point,
  title={Point-e: A system for generating 3d point clouds from complex prompts},
  author={Nichol, Alex and Jun, Heewoo and Dhariwal, Prafulla and Mishkin, Pamela and Chen, Mark},
  journal={arXiv preprint arXiv:2212.08751},
  year={2022}
}

@article{jun2023shap,
  title={Shap-e: Generating conditional 3d implicit functions},
  author={Jun, Heewoo and Nichol, Alex},
  journal={arXiv preprint arXiv:2305.02463},
  year={2023}
}

@article{zhang20233dshape2vecset,
  title={3dshape2vecset: A 3d shape representation for neural fields and generative diffusion models},
  author={Zhang, Biao and Tang, Jiapeng and Niessner, Matthias and Wonka, Peter},
  journal={ACM Transactions On Graphics (TOG)},
  volume={42},
  number={4},
  pages={1--16},
  year={2023},
  publisher={ACM New York, NY, USA}
}

@article{zhang2024clay,
  title={Clay: A controllable large-scale generative model for creating high-quality 3d assets},
  author={Zhang, Longwen and Wang, Ziyu and Zhang, Qixuan and Qiu, Qiwei and Pang, Anqi and Jiang, Haoran and Yang, Wei and Xu, Lan and Yu, Jingyi},
  journal={ACM Transactions on Graphics (TOG)},
  volume={43},
  number={4},
  pages={1--20},
  year={2024},
  publisher={ACM New York, NY, USA}
}

@inproceedings{xiang2025structured,
  title={Structured 3d latents for scalable and versatile 3d generation},
  author={Xiang, Jianfeng and Lv, Zelong and Xu, Sicheng and Deng, Yu and Wang, Ruicheng and Zhang, Bowen and Chen, Dong and Tong, Xin and Yang, Jiaolong},
  booktitle={Proceedings of the IEEE/CVF conference on computer vision and pattern recognition},
  pages={21469--21480},
  year={2025}
}

@article{li2025triposg,
  title={TripoSG: High-Fidelity 3D Shape Synthesis using Large-Scale Rectified Flow Models},
  author={Li, Yangguang and Zou, Zi-Xin and Liu, Zexiang and Wang, Dehu and Liang, Yuan and Yu, Zhipeng and Liu, Xingchao and Guo, Yuan-Chen and Liang, Ding and Ouyang, Wanli and others},
  journal={arXiv preprint arXiv:2502.06608},
  year={2025}
}

@misc{hunyuan3d2025hunyuan3d21,
    title={Hunyuan3D 2.1: From Images to High-Fidelity 3D Assets with Production-Ready PBR Material},
    author={Tencent Hunyuan3D Team},
    year={2025},
    eprint={2506.15442},
    archivePrefix={arXiv},
    primaryClass={cs.CV}
}

@inproceedings{tang2024lgm,
  title={Lgm: Large multi-view gaussian model for high-resolution 3d content creation},
  author={Tang, Jiaxiang and Chen, Zhaoxi and Chen, Xiaokang and Wang, Tengfei and Zeng, Gang and Liu, Ziwei},
  booktitle={European Conference on Computer Vision},
  pages={1--18},
  year={2024},
  organization={Springer}
}

@inproceedings{xu2024grm,
  title={Grm: Large gaussian reconstruction model for efficient 3d reconstruction and generation},
  author={Xu, Yinghao and Shi, Zifan and Yifan, Wang and Chen, Hansheng and Yang, Ceyuan and Peng, Sida and Shen, Yujun and Wetzstein, Gordon},
  booktitle={European Conference on Computer Vision},
  pages={1--20},
  year={2024},
  organization={Springer}
}

@inproceedings{hong2024lrm,
  title={Lrm: Large reconstruction model for single image to 3d},
  author={Hong, Yicong and Zhang, Kai and Gu, Jiuxiang and Bi, Sai and Zhou, Yang and Liu, Difan and Liu, Feng and Sunkavalli, Kalyan and Bui, Trung and Tan, Hao},
  booktitle={International Conference on Learning Representations},
  volume={2024},
  pages={50678--50702},
  year={2024}
}

@inproceedings{szymanowicz2024splatter,
  title={Splatter image: Ultra-fast single-view 3d reconstruction},
  author={Szymanowicz, Stanislaw and Rupprecht, Chrisitian and Vedaldi, Andrea},
  booktitle={Proceedings of the IEEE/CVF conference on computer vision and pattern recognition},
  pages={10208--10217},
  year={2024}
}

@inproceedings{ye2025hi3dgen,
  title={Hi3dgen: High-fidelity 3d geometry generation from images via normal bridging},
  author={Ye, Chongjie and Wu, Yushuang and Lu, Ziteng and Chang, Jiahao and Guo, Xiaoyang and Zhou, Jiaqing and Zhao, Hao and Han, Xiaoguang},
  booktitle={Proceedings of the IEEE/CVF International Conference on Computer Vision},
  pages={25050--25061},
  year={2025}
}

@article{li2026sparc3d,
  title={Sparc3d: Sparse representation and construction for high-resolution 3d shapes modeling},
  author={Li, Zhihao and Wang, Yufei and Zheng, Heliang and Luo, Yihao and Wen, Bihan},
  journal={Advances in Neural Information Processing Systems},
  volume={38},
  pages={118582--118600},
  year={2026}
}

@inproceedings{liu2024part123,
  title={Part123: Part-aware 3D reconstruction from a single-view image},
  author={Liu, Anran and Lin, Cheng and Liu, Yuan and Long, Xiaoxiao and Dou, Zhiyang and Guo, Hao-Xiang and Luo, Ping and Wang, Wenping},
  booktitle={ACM SIGGRAPH 2024 Conference Papers},
  pages={1--12},
  year={2024}
}

@inproceedings{chen2025partgen,
  title={Partgen: Part-level 3d generation and reconstruction with multi-view diffusion models},
  author={Chen, Minghao and Shapovalov, Roman and Laina, Iro and Monnier, Tom and Wang, Jianyuan and Novotny, David and Vedaldi, Andrea},
  booktitle={Proceedings of the Computer Vision and Pattern Recognition Conference},
  pages={5881--5892},
  year={2025}
}

@inproceedings{yang2025omnipart,
  title={Omnipart: Part-aware 3d generation with semantic decoupling and structural cohesion},
  author={Yang, Yunhan and Zhou, Yufan and Guo, Yuan-Chen and Zou, Zi-Xin and Huang, Yukun and Liu, Ying-Tian and Xu, Hao and Liang, Ding and Cao, Yan-Pei and Liu, Xihui},
  booktitle={Proceedings of the SIGGRAPH Asia 2025 Conference Papers},
  pages={1--12},
  year={2025}
}

@article{zhang2025bang,
  title={BANG: Dividing 3D assets via generative exploded dynamics},
  author={Zhang, Longwen and Zhang, Qixuan and Jiang, Haoran and Bai, Yinuo and Yang, Wei and Xu, Lan and Yu, Jingyi},
  journal={ACM Transactions on Graphics (TOG)},
  volume={44},
  number={4},
  pages={1--21},
  year={2025},
  publisher={ACM New York, NY, USA}
}

@article{lin2026partcrafter,
  title={Partcrafter: Structured 3d mesh generation via compositional latent diffusion transformers},
  author={Lin, Yuchen and Lin, Chenguo and Pan, Panwang and Yan, Honglei and Yiqiang, Feng and Mu, Yadong and Fragkiadaki, Katerina},
  journal={Advances in neural information processing systems},
  volume={38},
  pages={35387--35415},
  year={2026}
}

@article{tang2026partpacker,
  title={Efficient part-level 3d object generation via dual volume packing},
  author={Tang, Jiaxiang and Lu, Ruijie and Li, Max and Hao, Zekun and Li, Xuan and Wei, Fangyin and Song, Shuran and Zeng, Gang and Liu, Ming-Yu and Lin, Tsung-Yi},
  journal={Advances in Neural Information Processing Systems},
  volume={38},
  pages={27115--27137},
  year={2026}
}

@inproceedings{yan2024frankenstein,
  title={Frankenstein: Generating semantic-compositional 3d scenes in one tri-plane},
  author={Yan, Han and Li, Yang and Wu, Zhennan and Chen, Shenzhou and Sun, Weixuan and Shang, Taizhang and Liu, Weizhe and Chen, Tian and Dai, Xiaqiang and Ma, Chao and others},
  booktitle={SIGGRAPH Asia 2024 Conference Papers},
  pages={1--11},
  year={2024}
}

@article{chen2026autopartgen,
  title={AutoPartGen: Autoregressive 3D Part Generation and Discovery},
  author={Chen, Minghao and Wang, Jianyuan and Shapovalov, Roman and Monnier, Tom and Jung, Hyunyoung and Wang, Dilin and Ranjan, Rakesh and Laina, Iro and Vedaldi, Andrea},
  journal={Advances in Neural Information Processing Systems},
  volume={38},
  pages={153496--153521},
  year={2026}
}

@article{he2025unipart,
  title={UniPart: Part-Level 3D Generation with Unified 3D Geom-Seg Latents},
  author={He, Xufan and Wu, Yushuang and Guo, Xiaoyang and Ye, Chongjie and Zhou, Jiaqing and Hu, Tianlei and Han, Xiaoguang and Du, Dong},
  journal={arXiv preprint arXiv:2512.09435},
  year={2025}
}

@inproceedings{qi2017pointnet,
  title={Pointnet: Deep learning on point sets for 3d classification and segmentation},
  author={Qi, Charles R and Su, Hao and Mo, Kaichun and Guibas, Leonidas J},
  booktitle={Proceedings of the IEEE conference on computer vision and pattern recognition},
  pages={652--660},
  year={2017}
}

@inproceedings{qi2017pointnet++,
  title={PointNet++ deep hierarchical feature learning on point sets in a metric space},
  author={Qi, Charles R and Yi, Li and Su, Hao and Guibas, Leonidas J},
  booktitle={Proceedings of the 31st International Conference on Neural Information Processing Systems},
  pages={5105--5114},
  year={2017}
}

@inproceedings{zhao2021point,
  title={Point transformer},
  author={Zhao, Hengshuang and Jiang, Li and Jia, Jiaya and Torr, Philip HS and Koltun, Vladlen},
  booktitle={Proceedings of the IEEE/CVF international conference on computer vision},
  pages={16259--16268},
  year={2021}
}

@inproceedings{kirillov2023sam,
  title={Segment anything},
  author={Kirillov, Alexander and Mintun, Eric and Ravi, Nikhila and Mao, Hanzi and Rolland, Chloe and Gustafson, Laura and Xiao, Tete and Whitehead, Spencer and Berg, Alexander C and Lo, Wan-Yen and others},
  booktitle={Proceedings of the IEEE/CVF international conference on computer vision},
  pages={4015--4026},
  year={2023}
}

@inproceedings{liu2023partslip,
  title={Partslip: Low-shot part segmentation for 3d point clouds via pretrained image-language models},
  author={Liu, Minghua and Zhu, Yinhao and Cai, Hong and Han, Shizhong and Ling, Zhan and Porikli, Fatih and Su, Hao},
  booktitle={Proceedings of the IEEE/CVF conference on computer vision and pattern recognition},
  pages={21736--21746},
  year={2023}
}

@article{cen2023segment,
  title={Segment anything in 3d with nerfs},
  author={Cen, Jiazhong and Zhou, Zanwei and Fang, Jiemin and Shen, Wei and Xie, Lingxi and Jiang, Dongsheng and Zhang, Xiaopeng and Tian, Qi and others},
  journal={Advances in Neural Information Processing Systems},
  volume={36},
  pages={25971--25990},
  year={2023}
}

@inproceedings{abdelreheem2023satr,
  title={Satr: Zero-shot semantic segmentation of 3d shapes},
  author={Abdelreheem, Ahmed and Skorokhodov, Ivan and Ovsjanikov, Maks and Wonka, Peter},
  booktitle={Proceedings of the IEEE/CVF International Conference on Computer Vision},
  pages={15166--15179},
  year={2023}
}

@article{yang2023sam3d,
  title={Sam3d: Segment anything in 3d scenes},
  author={Yang, Yunhan and Wu, Xiaoyang and He, Tong and Zhao, Hengshuang and Liu, Xihui},
  journal={arXiv preprint arXiv:2306.03908},
  year={2023}
}

@article{chen2025sam,
  title={Sam 3d: 3dfy anything in images},
  author={Chen, Xingyu and Chu, Fu-Jen and Gleize, Pierre and Liang, Kevin J and Sax, Alexander and Tang, Hao and Wang, Weiyao and Guo, Michelle and Hardin, Thibaut and Li, Xiang and others},
  journal={arXiv preprint arXiv:2511.16624},
  year={2025}
}

@inproceedings{liu2025partfield,
  title={Partfield: Learning 3d feature fields for part segmentation and beyond},
  author={Liu, Minghua and Uy, Mikaela Angelina and Xiang, Donglai and Su, Hao and Fidler, Sanja and Sharp, Nicholas and Gao, Jun},
  booktitle={Proceedings of the IEEE/CVF International Conference on Computer Vision},
  pages={9704--9715},
  year={2025}
}

@article{yang2024sampart3d,
  title={Sampart3d: Segment any part in 3d objects},
  author={Yang, Yunhan and Huang, Yukun and Guo, Yuan-Chen and Lu, Liangjun and Wu, Xiaoyang and Lam, Edmund Y and Cao, Yan-Pei and Liu, Xihui},
  journal={arXiv preprint arXiv:2411.07184},
  year={2024}
}

@article{ma2025p3,
  title={P3-sam: Native 3d part segmentation},
  author={Ma, Changfeng and Li, Yang and Yan, Xinhao and Xu, Jiachen and Yang, Yunhan and Wang, Chunshi and Zhao, Zibo and Guo, Yanwen and Chen, Zhuo and Guo, Chunchao},
  journal={arXiv preprint arXiv:2509.06784},
  year={2025}
}

@article{li2026segvigen,
  title={SegviGen: Repurposing 3D Generative Model for Part Segmentation},
  author={Li, Lin and Feng, Haoran and Huang, Zehuan and Chen, Haohua and Nie, Wenbo and Hou, Shaohua and Fan, Keqing and Hu, Pan and Wang, Sheng and Li, Buyu and others},
  journal={arXiv preprint arXiv:2603.16869},
  year={2026}
}

@inproceedings{liu2023flow,
  title={Flow Straight and Fast: Learning to Generate and Transfer Data with Rectified Flow},
  author={Liu, Xingchao and Gong, Chengyue and Liu, Qiang},
  booktitle={The Eleventh International Conference on Learning Representations (ICLR)},
  year={2023}
}

@inproceedings{ravi2025sam,
  title={Sam 2: Segment anything in images and videos},
  author={Ravi, Nikhila and Gabeur, Valentin and Hu, Yuan-Ting and Hu, Ronghang and Ryali, Chaitanya and Ma, Tengyu and Khedr, Haitham and R{\"a}dle, Roman and Rolland, Chloe and Gustafson, Laura and others},
  booktitle={International Conference on Learning Representations},
  volume={2025},
  pages={28085--28128},
  year={2025}
}

@article{oquab2024dinov2,
  title={DINOv2: Learning Robust Visual Features without Supervision},
  author={Oquab, Maxime and Darcet, Timoth{\'e}e and Moutakanni, Th{\'e}o and Vo, Huy and Szafraniec, Marc and Khalidov, Vasil and Fernandez, Pierre and Haziza, Daniel and Massa, Francisco and El-Nouby, Alaaeldin and others},
  journal={Transactions on Machine Learning Research Journal},
  pages={1--31},
  year={2024}
}

@inproceedings{chetverikov2002trimmed,
  title={The trimmed iterative closest point algorithm},
  author={Chetverikov, Dmitry and Svirko, Dmitry and Stepanov, Dmitry and Krsek, Pavel},
  booktitle={2002 International Conference on Pattern Recognition},
  volume={3},
  pages={545--548},
  year={2002},
  organization={IEEE}
}

@inproceedings{peebles2023scalable,
  title={Scalable diffusion models with transformers},
  author={Peebles, William and Xie, Saining},
  booktitle={Proceedings of the IEEE/CVF international conference on computer vision},
  pages={4195--4205},
  year={2023}
}

@inproceedings{deitke2023objaverse,
  title={Objaverse: A universe of annotated 3d objects},
  author={Deitke, Matt and Schwenk, Dustin and Salvador, Jordi and Weihs, Luca and Michel, Oscar and VanderBilt, Eli and Schmidt, Ludwig and Ehsani, Kiana and Kembhavi, Aniruddha and Farhadi, Ali},
  booktitle={Proceedings of the IEEE/CVF conference on computer vision and pattern recognition},
  pages={13142--13153},
  year={2023}
}

@article{deitke2023objaversexl,
  title={Objaverse-xl: A universe of 10m+ 3d objects},
  author={Deitke, Matt and Liu, Ruoshi and Wallingford, Matthew and Ngo, Huong and Michel, Oscar and Kusupati, Aditya and Fan, Alan and Laforte, Christian and Voleti, Vikram and Gadre, Samir Yitzhak and others},
  journal={Advances in Neural Information Processing Systems},
  volume={36},
  pages={35799--35813},
  year={2023}
}

@inproceedings{collins2022abo,
  title={Abo: Dataset and benchmarks for real-world 3d object understanding},
  author={Collins, Jasmine and Goel, Shubham and Deng, Kenan and Luthra, Achleshwar and Xu, Leon and Gundogdu, Erhan and Zhang, Xi and Vicente, Tomas F Yago and Dideriksen, Thomas and Arora, Himanshu and others},
  booktitle={Proceedings of the IEEE/CVF conference on computer vision and pattern recognition},
  pages={21126--21136},
  year={2022}
}

@article{fu20213d,
  title={3d-future: 3d furniture shape with texture},
  author={Fu, Huan and Jia, Rongfei and Gao, Lin and Gong, Mingming and Zhao, Binqiang and Maybank, Steve and Tao, Dacheng},
  journal={International Journal of Computer Vision},
  volume={129},
  number={12},
  pages={3313--3337},
  year={2021},
  publisher={Springer}
}

@inproceedings{khanna2024habitat,
  title={Habitat synthetic scenes dataset (hssd-200): An analysis of 3d scene scale and realism tradeoffs for objectgoal navigation},
  author={Khanna, Mukul and Mao, Yongsen and Jiang, Hanxiao and Haresh, Sanjay and Shacklett, Brennan and Batra, Dhruv and Clegg, Alexander and Undersander, Eric and Chang, Angel X and Savva, Manolis},
  booktitle={Proceedings of the IEEE/CVF Conference on Computer Vision and Pattern Recognition},
  pages={16384--16393},
  year={2024}
}

@article{yang2025holopart,
  title={Holopart: Generative 3d part amodal segmentation},
  author={Yang, Yunhan and Guo, Yuan-Chen and Huang, Yukun and Zou, Zi-Xin and Yu, Zhipeng and Li, Yangguang and Cao, Yan-Pei and Liu, Xihui},
  journal={arXiv preprint arXiv:2504.07943},
  year={2025}
}

@article{yan2025x,
  title={X-part: high fidelity and structure coherent shape decomposition},
  author={Yan, Xinhao and Xu, Jiachen and Li, Yang and Ma, Changfeng and Yang, Yunhan and Wang, Chunshi and Zhao, Zibo and Lai, Zeqiang and Zhao, Yunfei and Chen, Zhuo and others},
  journal={arXiv preprint arXiv:2509.08643},
  year={2025}
}
